\documentclass[letterpaper, 10 pt, conference]{ieeeconf}
\IEEEoverridecommandlockouts
\let\labelindent\relax
\let\labelindent\relax
\usepackage{graphicx} \graphicspath{ {figures/} }
\usepackage{amsmath,amssymb,mathabx,amsbsy,mathtools,etoolbox}
\usepackage[utf8]{inputenc} 
\usepackage[T1]{fontenc}    
\usepackage{url}
\usepackage{algorithmic}
\usepackage[ruled]{algorithm2e}

\usepackage{acronym}
\usepackage{enumitem}
\usepackage{balance}
\usepackage{xspace}
\usepackage{setspace}
\usepackage[skip=3pt,font=small]{subcaption}
\usepackage[skip=3pt,font=small]{caption}
\usepackage[dvipsnames,svgnames,x11names]{xcolor}
\usepackage[capitalise,nameinlink]{cleveref}
\usepackage{booktabs,tabularx,colortbl,multirow,multicol,array,makecell}
\usepackage{pifont}
\usepackage{cite}
\usepackage{nicematrix}
\usepackage{dblfloatfix}

\makeatletter
\DeclareRobustCommand\onedot{\futurelet\@let@token\@onedot}
\def\@onedot{\ifx\@let@token.\else.\null\fi\xspace}

\def\etal{\emph{et al}\onedot}
\makeatother

\makeatletter
\def\BState{\State\hskip-\ALG@thistlm}
\makeatother

\makeatletter
\renewcommand{\paragraph}{%
  \@startsection{paragraph}{4}%
  {\z@}{0ex \@plus 0ex \@minus 0ex}{-1em}%
  {\hskip\parindent\normalfont\normalsize\bfseries}%
}
\makeatother

\crefname{algorithm}{Alg.}{Algs.}
\Crefname{algocf}{Algorithm}{Algorithms}
\crefname{section}{Sec.}{Secs.}
\Crefname{section}{Section}{Sections}
\crefname{table}{Tab.}{Tabs.}
\Crefname{table}{Table}{Tables}
\crefname{figure}{Fig.}{Fig.}
\Crefname{figure}{Figure}{Figure}

\definecolor{gblue}{HTML}{4285F4}
\definecolor{gred}{HTML}{DB4437}
\definecolor{ggreen}{HTML}{0F9D58}

\definecolor{mygray}{gray}{.92}

\definecolor{lightgray}{gray}{0.9}

\acrodef{vbts}[VBTS]{Vision-based Tactile Sensor}
\acrodef{qp}[QP]{Quadratic Programming}
\acrodef{dof}[DoF]{Degree of Freedom}
\acrodef{ros}[ROS]{Robot Operating System}
\acrodef{uav}[UAV]{Unmanned Aerial Vehicle}
\acrodef{dof}[DoF]{Degree of Freedom}
\acrodef{com}[CoM]{center of mass}
\acrodef{msrr}[MSRR]{modular self-reconfigurable robot}
\acrodef{owmr}[OWMR]{Omnidirectional Wheeled Mobile Robot}
\acrodef{wmr}[WMR]{Wheeled Mobile Robot}
\acrodef{ftac}[F-TAC Hand]{Full-hand TACtile-embedded anthropomorphic Hand}
\acrodef{gep}[GEP]{Generalized Earley Parser}
\acrodef{t-aog}[T-AOG]{temporal And-Or Graph}
\acrodef{dps}[DPS]{Deep Photometric Stereo}
\acrodef{mala}[MALA]{Metropolis-Adjusted Langevin Algorithm}
\acrodef{mcp}[MCP]{metacarpophalangeal}
\acrodef{pip}[PIP]{proximal interphalangeal}
\acrodef{dip}[DIP]{distal interphalangeal}
\acrodef{adelm}[ADELM]{Attraction-Diffusion Energy Landscape Mapping}
\acrodef{dof}[DoF]{Degree-of-Freedom}
\acrodef{meps}[MEPs]{Minimum Energy Paths}
\acrodef{pca}[PCA]{Principle Component Analysis}
\acrodef{svc}[SVC]{Support Vector Classifier}
\acrodef{fpsampling}[FPS]{Furthest Point Sampling}
\acrodef{t-SNE}[t-SNE]{t-distributed Stochastic Neighbour Embedding}
\acrodef{rbf}[RBF]{Radial Basis Function}

\title{\LARGE \bf 3D Vision-tactile Reconstruction from Infrared and Visible Images\\ for Robotic Fine-grained Tactile Perception}
\author{Yuankai Lin$^{1}$, Xiaofan Lu$^{1}$, Jiahui Chen$^{1}$, Hua Yang$^{1\dagger}$
\thanks{$\dagger$ Corresponding authors.}
\thanks{$^{1}$ All authors are with the State Key Laboratory of Digital Manufacturing Equipment and Technology, School of Mechanical Science and Engineering, Huazhong University of Science and Technology, Wuhan 430074, China. (Emails: huayang@hust.edu.cn)}
}

\begin{document}

\maketitle   

\begin{abstract}
To achieve human-like haptic perception in anthropomorphic grippers, the compliant sensing surfaces of vision tactile sensor (VTS) must evolve from conventional planar configurations to biomimetically curved topographies with continuous surface gradients. 
However, planar VTSs have challenges when extended to curved surfaces, including insufficient lighting of surfaces, blurring in reconstruction, and complex spatial boundary conditions for surface structures. 
With an end goal of constructing a human-like fingertip, our research (i) develops GelSplitter3D by expanding imaging channels with a prism and a near-infrared (NIR) camera, (ii) proposes a photometric stereo neural network with a CAD-based normal ground truth generation method to calibrate tactile geometry, and (iii) devises a normal integration method with boundary constraints of depth prior information to correcting the cumulative error of surface integrals. We demonstrate better tactile sensing performance, a 40$\%$ improvement in normal estimation accuracy, and the benefits of sensor shapes in grasping and manipulation tasks.
\end{abstract}

\section{Introduction}

Humans have shown remarkable adaptability to grasping and manipulating tasks, ranging from grasping tools under various conditions to fine manipulation of tiny objects~\cite{wu2023learning}. This capability, specifically the high-precision perception of tactile geometry, has fueled significant advances in robotic tactile sensing, which is now recognized as a pivotal enhancement for robotic perception systems~\cite{Sun2022,10563188}. The applications of these span a broad spectrum, including object detection~\cite{li2020skin}, six-degree-of-freedom (6-DOF) force estimation~\cite{sundaralingam2019robust}, robust grasping mechanisms~\cite{babin2021mechanisms}, gait planning strategies~\cite{wu2019tactile}, and imitation learning frameworks~\cite{fang2019survey}. These diverse applications necessitate the utilization of actuator ends with intricate surface geometries, equipped with highly accurate tactile sensors. Among all tactile sensors, the vision tactile sensor (VTS) has gained increasing adoption in robotic manipulation, due to its unique advantages, including high spatial resolution and multimodal sensing capabilities\cite{zhang2025artificial}. 

Despite these advancements, current VTSs predominantly adopt planar or simplistic curved configurations~\cite{romero2020soft, do2022densetact, yuan2017gelsight, gomes2020geltip, tippur2023gelsight360, zhang2024gelroller, tippur2024rainbowsight,10.1007/978-981-99-6498-7_2,lin20239dtact,xu2024dtactive}, which fundamentally limit their compatibility with anthropomorphic designs. Such geometric constraints create two critical barriers: (1) reduced contact area continuity when interacting with complex object geometries, leading to incomplete tactile feedback; (2) inherent photometric distortion in curved sensor surfaces that undermines existing photometric stereo algorithms calibrated for planar geometries. Recent attempts to address these limitations through multi-camera arrays~\cite{liu2023gelsight} or depth sensor fusion~\cite{alspach2019soft} have introduced new challenges in system miniaturization and computational complexity.

\begin{figure}[t!]
    \centering
    \includegraphics[width=\linewidth,trim=0cm 0cm 0cm 0cm, clip]{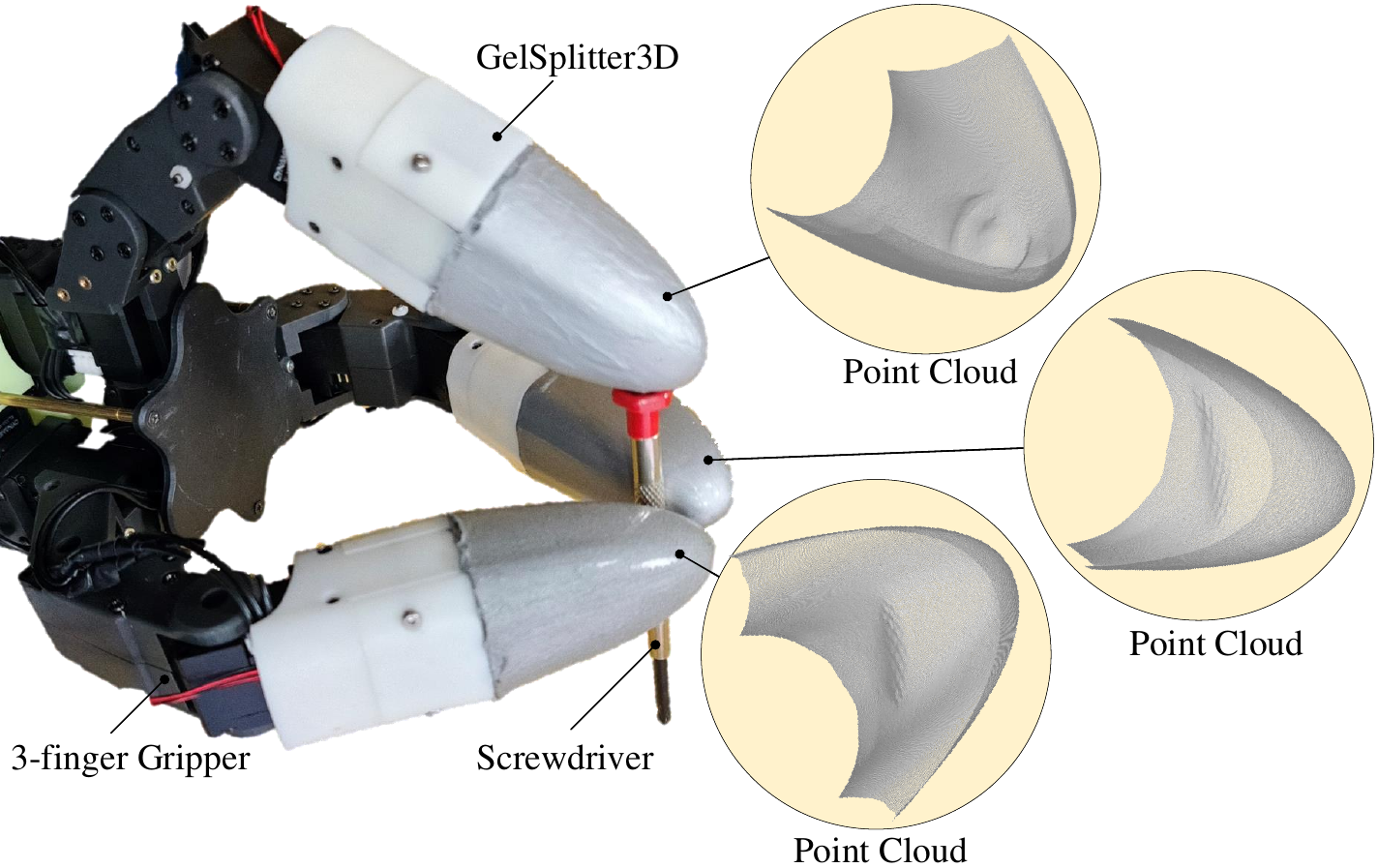}
    \caption{\textbf{GelSplitter3D sensors are deployed on a 12-Dof 3-finger gripper for robotic tactile fine-grained perception.} Full-frame images is reconstructed into dense 3D point clouds while preserving pixel-level geometric details. }
    \label{fig0}
\end{figure}

To bridge this gap, we present GelSplitter3D – a curved VTS system that provides biomimetic geometry with computational perception, as shown in Fig. \ref{fig0}. 
Unlike planar systems, curved surfaces exhibit large gradient variations. It is complex to achieve satisfactory imaging with three channels. The missing illumination information results in ambiguity and blurring in the reconstruction. Therefore, we propose RGB-NIR imaging to obtain more modal observations. Furthermore, a photometric stereo neural network (PSNN) is employed to estimate surface normals with a CAD-driven calibration process. Finally, to ensure geometric consistency, we incorporated boundary constraints based on depth priors. Our framework achieves 3D reconstruction of surfaces without the need for mathematical analysis.
To improve the performance in curved tactile sensing, our contributions are threefold:
\begin{enumerate}[]
\item 
The optical imaging bottleneck is addressed through prism-mediated multichannel light-field redirection captured by an additional near-infrared (NIR) camera.
\item
An PSNN architecture is established using RGB-NIR input and trained with CAD-simulated photometric patterns, enabling accurate surface normal estimation with a 40$\%$ decrease in MAE error. 
\item
A normal integration framework is developed that incorporates boundary constraints derived from prior sensor geometry, effectively reducing the propagation of cumulative errors in surface reconstruction.
\end{enumerate}

Qualitative and quantitative experiments validate the effectiveness of our GelSplitter3D sensor and the proposed method in the reconstruction of the curve surface.

\subsection{Related Work}

\textbf{Curved Surface VTSs:} VTSs excel in high-resolution multimodal fusion for robotic manipulation. Recent advances in curved-surface VTS designs have demonstrated significant advantages over planar configurations, particularly in expanding sensing coverage and enhancing anthropomorphic compatibility.  Romero \etal \cite{romero2020soft} pioneered light piping architectures that allow directional illumination control on curved surfaces. Liu \etal \cite{liu2023gelsight} engineered fingertip-shaped sensors with two cameras. Zhao \etal \cite{zhao2023gelsight} introduced flexible backbone deformation models for torque estimation. Tippur \etal \cite{tippur2023gelsight360} further optimized compatibility with varied curved geometries through cross-LED illumination. GelStereo Palm~\cite{10035880} proposes a curved visuotactile sensor, which senses the 3D contact geometry using a binocular vision system. Gelstereo BioTip~\cite{10358360} presents a soft, biomimetic fingertip-sized visuotactile sensor, which can perceive 3D surface deformation. However, most of the curved surface vision-tactile sensors mentioned above have simple geometric shapes such as spheres or cones, which limits their sensing capability in fine manipulation tasks and generalization across different operational tasks.

In this research, we propose a human finger-shaped curved surface VTS, capable of sensing the geometry of contact objects over a wide range, even in the absence of a mathematical analytical expression for the sensor's surface, and able to replicate human actions such as pinching, grabbing, and gripping.

\textbf{Photometric Stereo Methods:} In previous research, most VTSs utilizing photometric stereo for 3D reconstruction were limited to planar contact scenarios. To achieve 3D reconstruction on curved contact surfaces, researchers have explored various approaches. Zhang \etal \cite{zhang2024gelroller} proposed a self-supervised deep learning-based photometric stereo method that obtains surface normals from single images without pre-calibration. Tippur \etal \cite{tippur2024rainbowsight} developed a rainbow illumination scheme using semi-specular coatings to generate mixed color gradients. Gomes \etal \cite{gomes2020geltip} derived a projection function for curved sensors to map image-space pixels to physical surface points. Sun \etal \cite{sun2022soft} integrated photometric stereo with structured light, employing collimators to detect 3D deformations in flexible shells. However, existing photometric stereo methods primarily rely on RGB triple-channel information, demonstrating limited performance in fine texture perception. To address this limitation, we propose an RGB-NIR fusion photometric stereo neural network that significantly improves 3D reconstruction accuracy. 

\textbf{Normal Integration:}  Current vision tactile 3D reconstruction predominantly employs a fast Poisson solver for depth recovery via normal integration, yet faces challenges of unknown boundary conditions and cumulative errors in curved surface reconstruction. Cao \etal \cite{cao2022bilateral} introduced a bilateral weighted functional for semi-smooth surfaces, enabling discontinuity-preserving normal integration under perspective projection. Xiu \etal \cite{xiu2023econ} addressed global depth drift through variational optimization that reconstructs high-frequency details while incorporating low-frequency depth priors. Taketomi \etal \cite{cao2024supernormal} employed volume rendering to optimize a neural signed distance function (SDF), ensuring consistency between rendered and input normal maps. Building on these advances, we propose a normal integration framework that integrates boundary constraints with depth priors, effectively eliminating cumulative errors in curved surface reconstruction. 

\begin{figure}[t!]
    \centering
    \includegraphics[width=0.7\linewidth,]{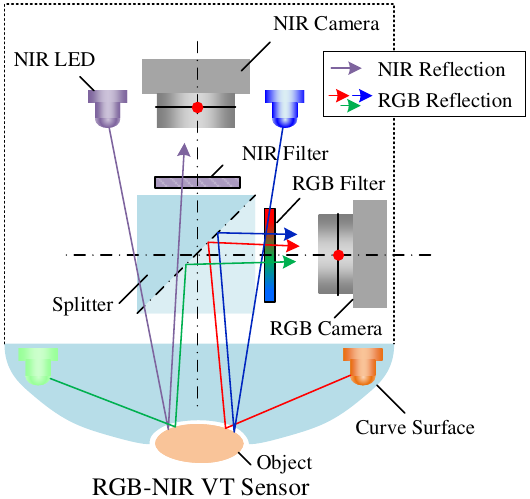}
    \caption{Compared to existing VTSs, our GelSplitter3D extends the dimensionality of perception while maintaining the same image plane. The bottleneck in optical imaging is addressed by employing prism-mediated multichannel light-field redirection, captured by an additional NIR camera.}
    \label{fig1}
\end{figure}

\begin{table}[t!]
    \caption{\textbf{The curve surface GelSight-like systems.} Our framework achieves 3D reconstruction of surfaces without the need for mathematical analysis.}
    \centering
    \label{tab:bom}
    \resizebox{0.95\linewidth}{!}{%
        \rowcolors{1}{white}{lightgray}
        \begin{tabular}{lccc}
        \toprule
        & \textbf{Name} & {\makecell[c]{\textbf{No. of sensors' channel}\\  \textbf{working simultaneously}}}& {\makecell[c]{\textbf{Requirement of}\\  \textbf{surface math analytic}}}\\
        \hline
        & Omnitact~\cite{padmanabha2020omnitact}& 3&  Yes\\
 & DenseTact~\cite{do2022densetact}& 3&Yes\\
 & GelRoller~\cite{zhang2024gelroller}& 3&Yes\\
 & GelTip~\cite{gomes2020geltip}& 3&Yes\\
 & GelSight360~\cite{tippur2023gelsight360}& 3&No\\
        & \textbf{Ours}& \textbf{6}& \textbf{No}\\
        \hline
        \end{tabular}
        }%
\end{table}

\subsection{Overview}
The remainder of this paper is organized as follows: Section II discusses related work in curved tactile sensing and photometric stereo. Section III details the GelSplitter3D hardware design and optical characterization. Section IV presents the neural network architecture and normal integration methodology. Section V provides experimental results, followed by conclusions in Section VI.

\section{Design \& Fabrication of GelSplitter3D}\label{sec:design}

\begin{figure}[t!]
\centering
\includegraphics[width=\linewidth]{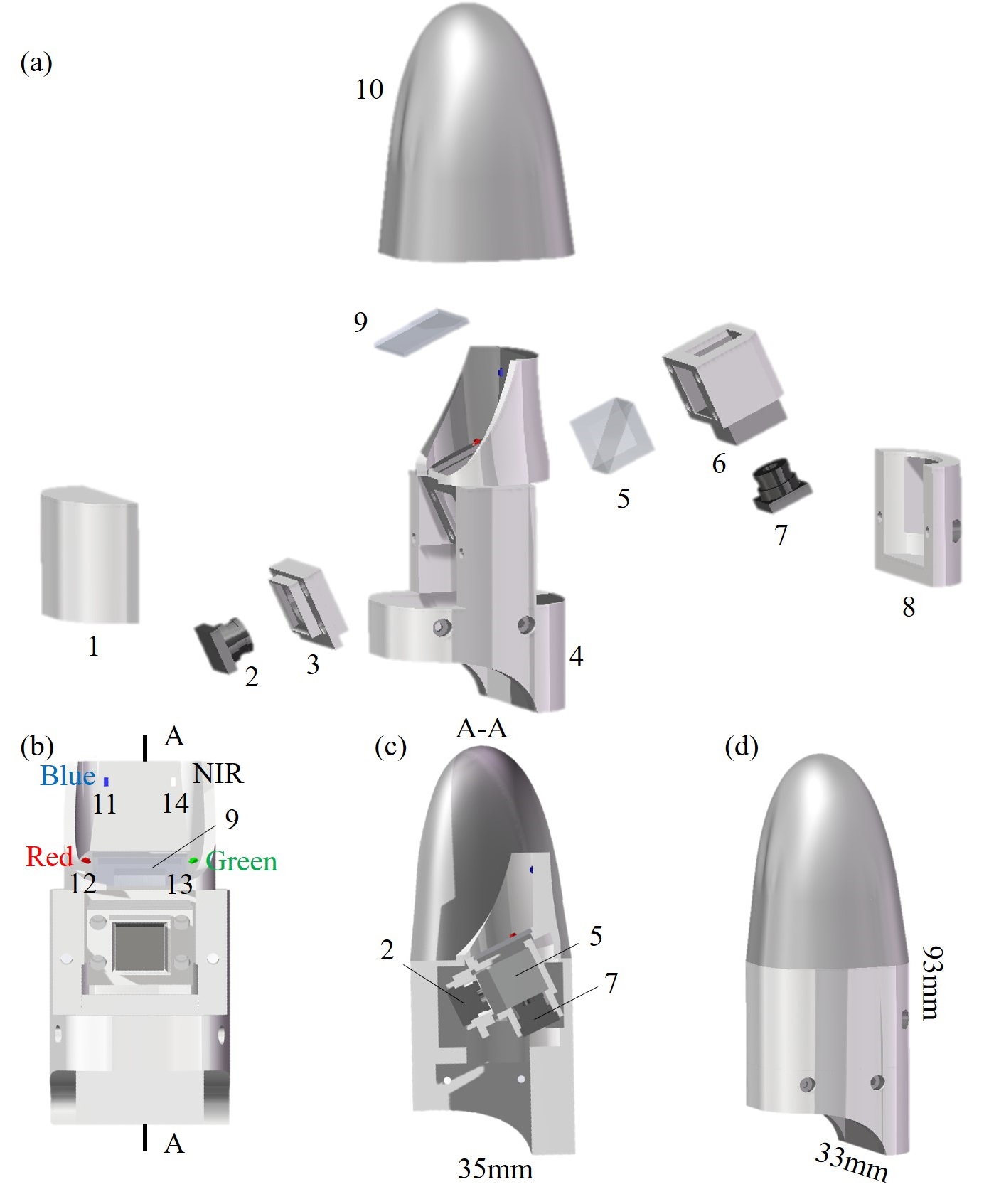}
\caption{\textbf{Design of GelSplitter3D sensor.}  1, 8: shells; 2, 7: RGB camera, NIR camera; 3, 6: camera mounts; 4: backbone; 5: prism; 9: acrylic; 10: gel contact module; 11, 12, 13, 14: RGB and NIR LEDs. (a) The exploded view of the sensor. (b) The lighting system consists of four built-in LEDs(RGB and NIR). (c) The cross-sectional view of the sensor's central plane. (d) The overall dimensions of the sensor are approximately $93 \times 35 \times 33~\mathrm{mm}$.}
\label{fig:fab}
\end{figure} 

\begin{figure}[t!]
\centering
\includegraphics[width=\linewidth]{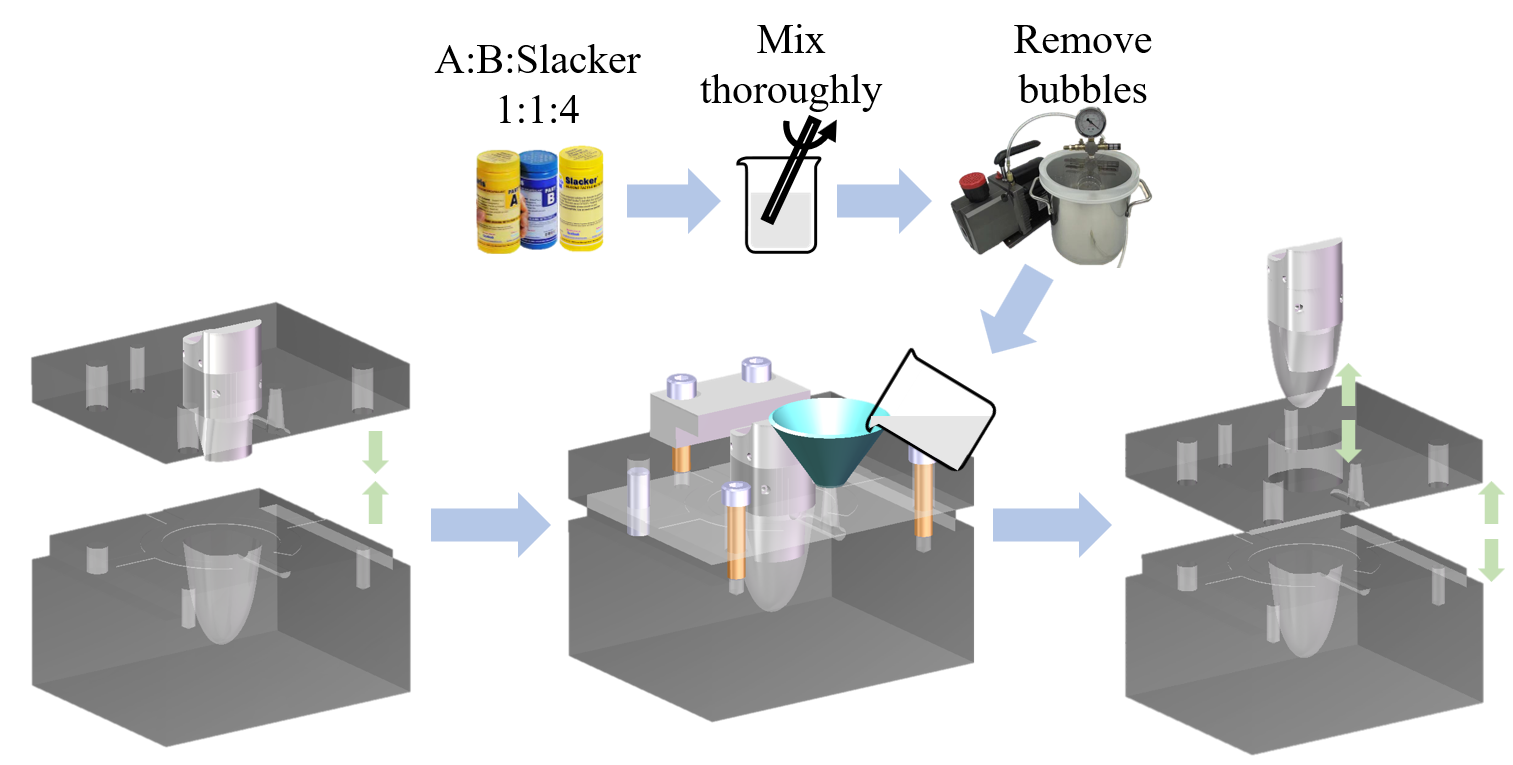}
\caption{\textbf{Design of the mold.} We adopt a top-bottom mold splitting approach to eliminate parting line marks on elastomer surfaces . }
\label{fig:ela}
\end{figure} 
The proposed sensor comprises two shells, two cameras and their respective camera mounts, four LEDs, a backbone, a prism, an acrylic, a gel contact module, as illustrated in Fig. \ref{fig:fab}(a). The overall dimensions of the sensor are approximately $93 \times 35 \times 33~\mathrm{mm}$, as depicted in Fig. \ref{fig:fab}(c)(d). The fabrication of each component is detailed below.

\textbf{Camera:} As illustrated in Fig. \ref{fig:fab}(a), an RGB camera and an NIR camera are used to capture images in the visible light and near-infrared bands, respectively. Both cameras used OmniVision's OV5640 CMOS sensors, a 650nm bandpass filter and a 940nm narrowband filter are integrated into the lens to separate the RGB and NIR components. To establish a controlled lighting environment, the cameras' auto white balance and auto exposure correction features are disabled. To achieve higher image quality, the camera resolution is set to $640 \times 480~\mathrm{mm}$, the camera lens is rotated to ensure the sensor surface deformation is within the depth of field, and the exposure compensation is adjusted to achieve appropriate brightness in the images.  

\textbf{Internal Illumination System:} The sensor's lighting system consists of four 0603-packaged surface-mount LEDs, including RGB and NIR. To ensure the accuracy of subsequent normal vector estimation using photometric stereo\cite{woodham1979photometric}, the positions of the four LEDs should not be too close to each other.

\textbf{Support and Fixation Structure:} The backbone, shells, and camera mounts are all made of photosensitive resin 9600 using 3D printing. This material offers advantages such as low cost, fine surface texture, good high-temperature resistance, and high toughness. Moreover, the backbone has a complex shape, making it suitable for manufacturing through 3D printing.

\textbf{Prism:} The prism is a cube with a side length of 10mm, a spectral ratio of 1:1, and a refractive index of 1.5168, capable of creating two identical fields of view. Except for the faces facing the gel and the two transparent cameras, all other faces are coated in black to reduce secondary reflections and the impact of external light.

\textbf{Elastomer:} To make the elastomer, firstly, assemble the backbone and shells, place them into the mold, and secure them. Then the A and B components of Solaris silicone (Smooth-on) are mixed with Slacker softener (Smooth-on) in a 1:1:4 weight ratio (Solaris A : Solaris B : Slacker), where a higher proportion of Slacker results in a softer silicone. After using a vacuum pump to remove bubbles from the liquid silicone, it is poured into the mold. After 24 hours of curing, separate the two mold parts to obtain the elastomer, as shown in Fig. \ref{fig:ela}. To avoid mold lines on the elastomer surface, we use a top-bottom mold splitting method, which ensures the cosmetic integrity of the silicone surface. To improve the smoothness of the surface, the lower mold is precision-machined from optical-grade acrylic (PMMA) instead of conventional 3D-printed molds that exhibit stair-stepping artifacts. 

\textbf{Reflective Paint}: The reflective paint, which is made by mixing silicone solvent, 800 mesh silver powder, and a thinner, is applied to the surface using a spray gun or brush. The silicone solvent ensures the adhesion of the reflective paint, the silver powder makes the surface resemble a Lambertian surface, and the thinner prevents the paint from curing too quickly.
\begin{figure*}[t]
    \centering
    \includegraphics[width=\textwidth]{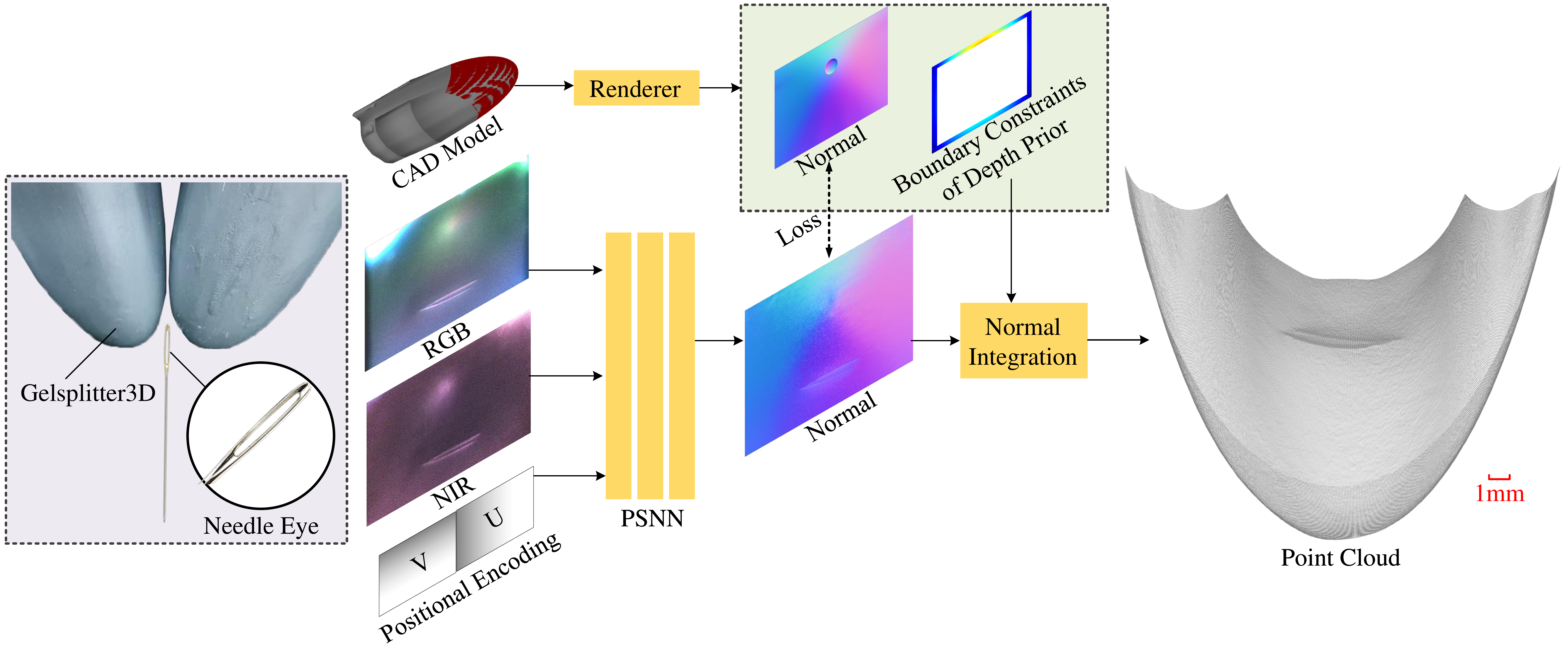}
    \caption{\textbf{The overview of our proposed PSNN and normal integration method with boundary constraints.} The proposed method enables accurate reconstruction of full-frame images into dense 3D point clouds while preserving pixel-level geometric details (e.g., needle eyes). }
    \label{fig3}
\end{figure*}

\section{Sensor Calibration of GelSplitter3D}\label{sec:cali_formation}
In this section, we describe a CAD-based calibration approach to the deployment of the GelSplitter3D proposed. We then present a normal integration method with boundary constraints to decrease the cumulative error of surface integrals.
\subsection{Geometry Calibration with CAD model}
\begin{figure}[t!]
    \centering
    \includegraphics[width=\linewidth]{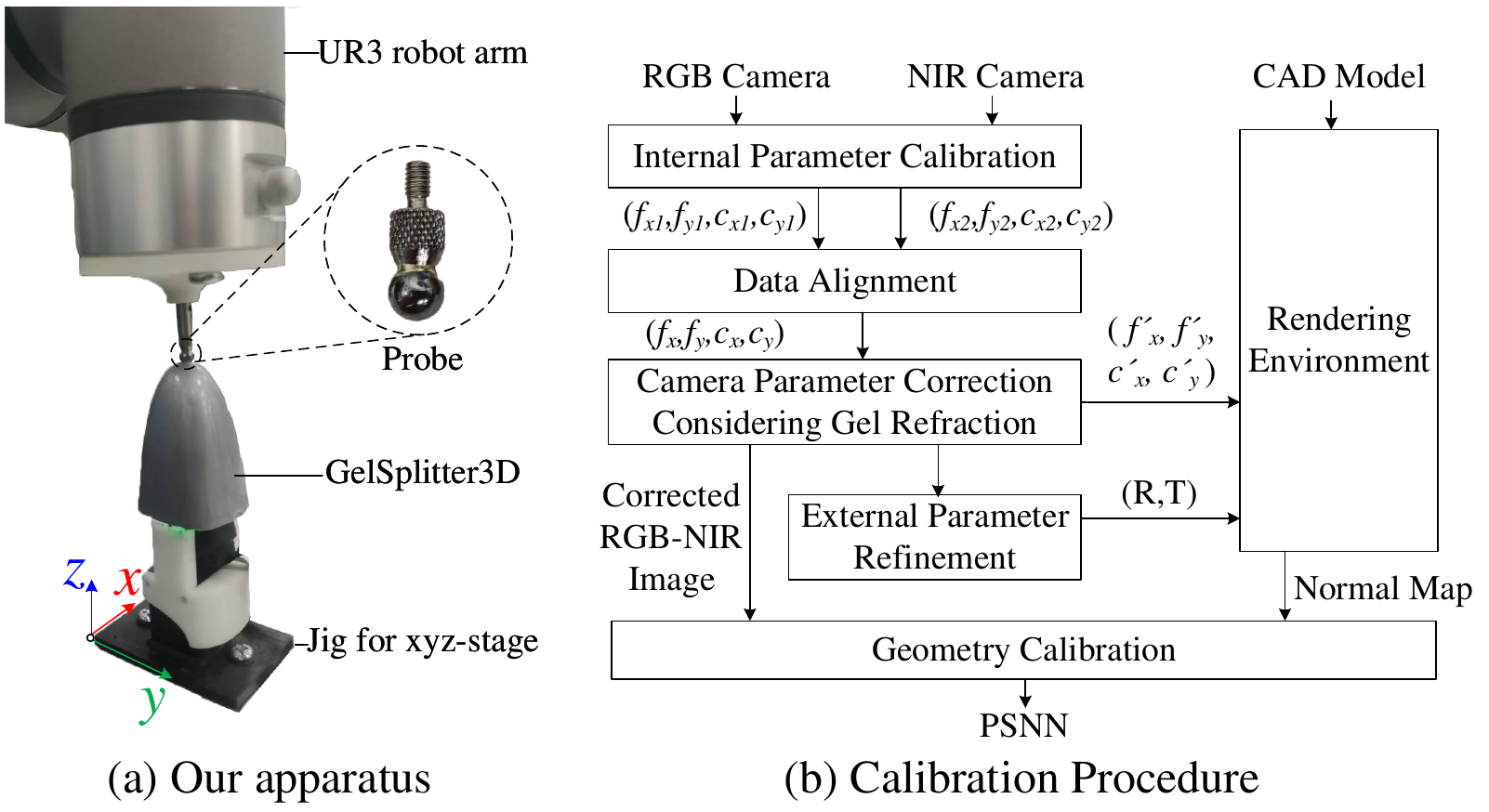}
    \caption{Our apparatus is used for geometry calibration. (a) A spherical probe with d = 5.0 mm is used. (b) A calibration procedure to obtain accurate truth values from a CAD model.}
    \label{fig2}
\end{figure}

A calibration procedure is designed to obtain accurate truth values from a CAD model, as shown in Fig. \ref{fig2}. First, The intrinsic parameters of both RGB and NIR cameras are initially calibrated using Zhang’s classical checkerboard method with sub-pixel corner detection. Then, random sample consensus (RANSAC) regression is used to align the checkerboard corners in the images of the two cameras for data alignment. To address complex light transport through the multi-medium optical system, a linear correction model is implemented to compensate for optical refraction caused by refractive interfaces, ensuring rendered viewpoints maintain spatial consistency with physical measurements:
\begin{equation}
f_{\text{medium}}=f_{\text{air}} \cdot \frac{n_{\text{medium}}}{n_{\text{air}}} . 
\end{equation}
Where $n_{\text{air}}=1.0$ is the refractive index of air and $n_{\text{medium}}=1.5168$ is the refractive index of the external medium. Manual fine-tuning of the parameters $(f_x,f_y,c_x,c_y)$ and $ \left[ R,T \right] $ is also performed to ensure that the simulated and actual fields of view are consistent. 

An apparatus for collecting the probed RGB-NIR image is shown in Fig. \ref{fig2} (a). The generation of ground truth serves two purposes: 1) to generate the normal maps of the sensor surface when probed with the calibration ball corresponding to each point of the pixel image and 2) to generate a binary mask of the ball contact region to collect the desired pixel coordinates and intensity values from the probed images.

In this study, our PSNN is implemented by a multiple-layer perception neural network (MLP) to learn the mapping between the intensity of the pixels in six channels, the positional encoding, and the normal map, as depicted in \cref{fig3}. The network is composed of three hidden layers, three dropout layers, and three batch normalization layers with the $relu$ activation function. In this paper, the PSNN first predicts accurate normal vectors $(n_x,n_y,n_z)$ from RGB-NIR images: \begin{equation}
(n_x,n_y,n_z) = \text{PSNN} (RGB,NIR). 
\end{equation}

The ground truth of normal vectors $(n_x',n_y',n_z')$ is rendered by the calibrated camera internal parameters $(f_x,f_y,c_x,c_y)$ and external parameters $ \left[ R,T \right] $. 

\begin{equation}
(n_x',n_y',n_z'), z_{\text{prior}} = \text{Renderer} (f_x,f_y,c_x,c_y,R,T). 
\end{equation}
Where the parameters $(f_x,f_y,c_x,c_y)$ and $ \left[ R,T \right] $ can be calculated by a calibration procedure, as shown in Fig. \ref{fig2} (b). $z_{\text{prior}}$ is the depth information of the 10-pixel-wide image edges, calculated from the CAD model. 

To further obtain the 3D shape, we adopt a normal integration method with boundary constraints for the curve surface.

\subsection{Normal Integration for Curve Surface}
Given a normal map $\mathbf{n}(x,y) = (n_x, n_y, n_z)$, the gradient field $(p, q)$ of the depth function $z(x,y)$ is computed as:
\begin{equation}
p = \frac{\partial z}{\partial x} = -\frac{n_x}{n_z }, \quad 
q = \frac{\partial z}{\partial y} = -\frac{n_y}{n_z}
\label{eq:gradient}
\end{equation}

The surface reconstruction problem can be formulated as\cite{kazhdan2006poisson}:
\begin{equation}
\nabla^2 z = \frac{\partial^2 z}{\partial x^2}+\frac{\partial^2 z}{\partial y^2}=\frac{\partial p}{\partial x} + \frac{\partial q}{\partial y}
\label{eq:poisson}
\end{equation}

We discretize \eqref{eq:poisson} using central differences. For valid pixels $(i,j)$:
\begin{align}
\frac{\partial^2 z}{\partial x^2} &\approx z_{i+1,j} + z_{i-1,j} - 2z_{i,j}\label{eq1} \\
\frac{\partial^2 z}{\partial y^2} &\approx z_{i,j+1} + z_{i,j-1} - 2z_{i,j} \label{eq2} \\\
\frac{\partial p}{\partial x} &\approx \frac{p_{i+1,j} - p_{i-1,j}}{2} \label{eq3} \\\
\frac{\partial q}{\partial y} &\approx \frac{q_{i,j+1} - q_{i,j-1}}{2}\label{eq4} \
\end{align}


For all valid pixels in the image, we assemble the corresponding Poisson equation \eqref{eq:poisson} into a sparse linear system: 
\begin{equation}
\mathbf{A}_{\mathrm{poisson}} \,\mathbf{z} = \mathbf{b}_{\mathrm{poisson}}
\label{eq:poisson_system}
\end{equation}
where \(\mathbf{A}_{\mathrm{poisson}}\) is the sparse matrix where each row encodes the coefficients of one pixel's equation, \(\mathbf{z}\) denotes the vector of all pixel depths, and \(\mathbf{b}_{\mathrm{poisson}}\) represents the divergence terms.

Incorporating depth priors $\mathbf{z}_{\text{prior}}$ with weight $\lambda$, the fused depth estimation can be formulated as an augmented linear system :
\begin{equation}
\begin{bmatrix}
\mathbf{A}_{\mathrm{poisson}} \\
\sqrt{\lambda}\mathbf{I}_{\text{prior}}
\end{bmatrix}
\mathbf{z} = 
\begin{bmatrix}
\mathbf{b}_{\mathrm{poisson}} \\
\sqrt{\lambda}\mathbf{z}_{\text{prior}}
\end{bmatrix}
\label{eq:augmented_system}
\end{equation}
Where \(\mathbf{I}_{\text{prior}}\) is a diagonal matrix with 1s at positions corresponding to pixels with valid depth priors. 

By introducing the prior of zero depth at the surface boundary, GelSplitter 3D ensures that all integration paths converge to the same reference plane at the boundary, thereby preventing the global divergence of errors caused by normal vector inaccuracies. In this paper, sensor edges are assumed to move by default, so depth information from the 10-pixel-wide image edges, derived from the CAD model, serves as a prior for normal integration.

\begin{table}[b!]
\small
    \caption{\textbf{Evaluation of proposed calibration methods on GelSplitter3D using look-up table and PSNN}}
    \centering
    \label{tab2}
    \resizebox{1\linewidth}{!}{%
        \begin{tabular}{c|cccc}
        \hline
        \multirow{3}{*}{\textbf{Groups}/(mm/pixel)} & \multicolumn{4}{c}{\textbf{Calibration Approaches}} \\ 
        \cline{2-5}
                  & \makecell[c]{Look-up Table~\cite{yuan2017gelsight} \\ RGB-only}
                  & \makecell[c]{Look-up Table~\cite{yuan2017gelsight} \\ RGB-NIR}
                  & \makecell[c]{PSNN \\ RGB-only} 
                  & \makecell[c]{PSNN \\ RGB-NIR} \\
        \hline
        Data Collected & $50$ & $50$  & $50$  &  $50$ \\
        \hline
        $G_x$ error (MAE) & $0.0389$ & $0.0359$  & $0.0108$  &  $\textbf{0.0072}$\color{red}$\downarrow 33\%$\\
        $G_y$ error (MAE) & $0.0316$ & $0.0286$  & $0.0108$  & $\textbf{0.0058}$\color{red}$\downarrow 46\%$  \\
        $Total$ error (MAE) & $0.0706$ & $0.0646$  & $0.0218$  & $\textbf{0.0130}$\color{red}$\downarrow 40\%$ \\
        \hline
        \end{tabular}
        }%
\end{table}

\begin{table}[b!]
\small
    \caption{\textbf{Evaluation of Curve Surface Normal Integration Methods Using GelSplitter3D.}}
    \centering
    \label{tab3}
    \resizebox{1\linewidth}{!}{%
        \begin{tabular}{c|ccc}
        \hline
        \multirow{3}{*}{\textbf{Groups}/(mm)} & \multicolumn{3}{c}{\textbf{Normal Integration Approaches}} \\ 
        \cline{2-4}
                  & \makecell[c]{Fast Poisson~\cite{yuan2017gelsight} \\ w/o Depth Prior}
                  & \makecell[c]{Ours \\ w/o Depth Prior}
                  & \makecell[c]{Ours \\ w/  Depth Prior}\\
        \hline
        Depth error (MAE) & $0.579$ & $0.1255$   &  $0.0406$\\
        \hline
        \end{tabular}
        }%
\end{table}

\section{RESULTS} 
In this section, qualitative and quantitative experiments are performed to validate the proposed PSNN method and the normal integration method.
\subsection{Normal Estimation by PSNN} 
The UR3 arm is used to ensure the stability of the probe spheres in contact with the gel and a $5.0$ mm sphere probe is used to uniformly sample the gel surface. 50 tactile images are collected on the apparatus. The sensor can be easily switched between RGB-only and RGB-NIR modes and provides a fair comparison of the results, as shown in Tab. \ref{tab2}. 

Quantitatively, our PSNN method is more suitable for normal integration in the context of curved surfaces. In terms of the gradients, $G_x$ and $G_y$, the error of the PSNN method decreased from 0.0389 and 0.0316 to 0.108 and 0.108, respectively, compared to the commonly used Look-up table method. This demonstrates the effectiveness of the proposed method in normal estimation under curved surface conditions. Furthermore, the NIR modality was incorporated into the PSNN, resulting in observed reductions in the errors of $G_x$ and $G_y$ to 0.0072 and 0.0058, respectively. The magnitude of this reduction is significant, amounting to an overall 40$\%$ decrease, which is crucial for enhancing the granularity of tactile perception.

\begin{figure*}[ht!]
    \centering
    \includegraphics[width=\linewidth]{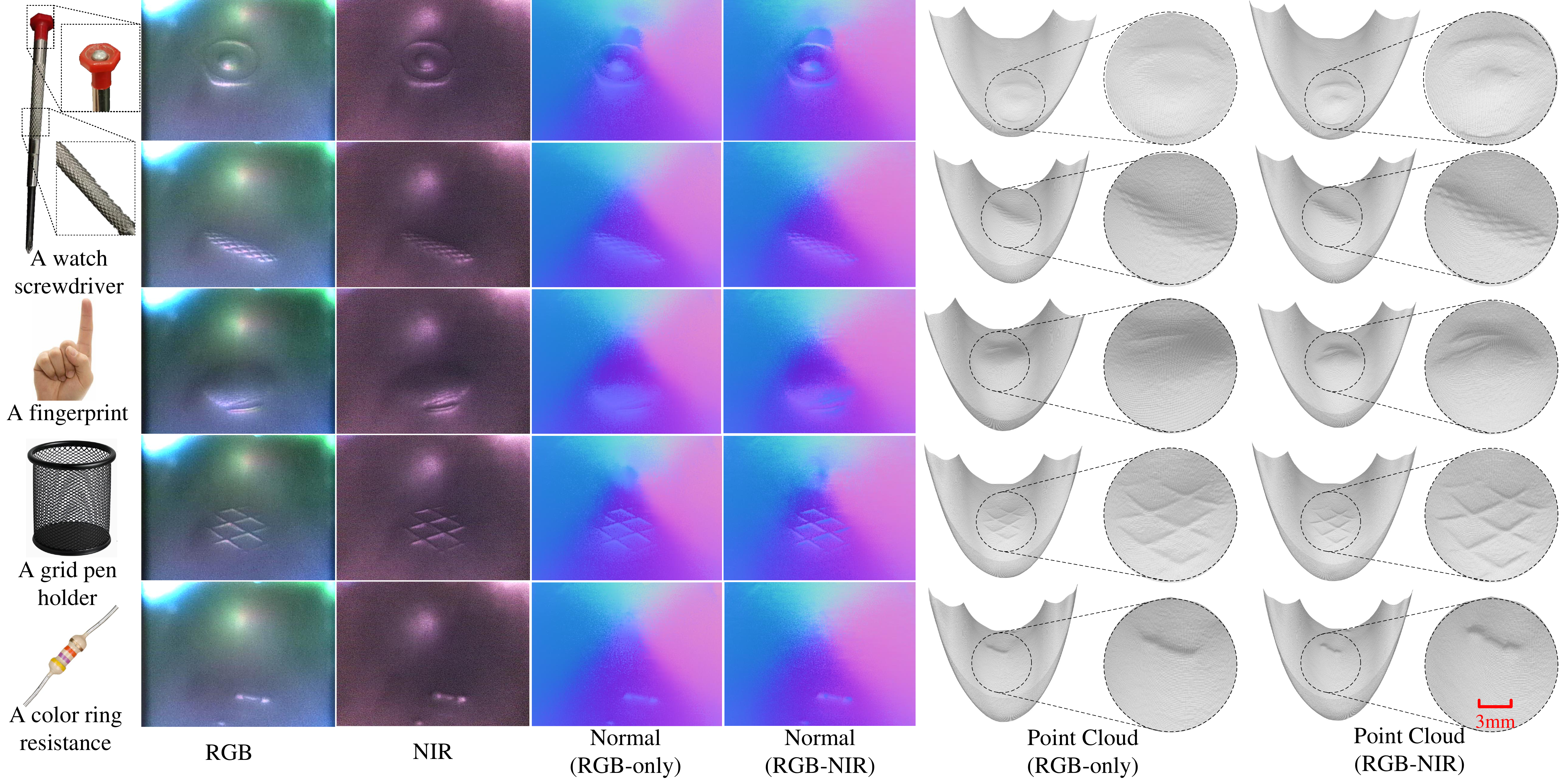}
    \caption{\textbf{Result of 3D tactile reconstruction from RGB-NIR.} Our GelSplitter can be easily switched between RGB and RGB-NIR modes to provide a fair comparison. We prepared items that require fine perception, such as watch screwdrivers, fingerprints, grids, and color ring resistors. Due to the complementary shape and texture information provided by NIR, higher clarity and fidelity performance are achieved.}
    \label{fig4}
\end{figure*}

Qualitatively, this improvement stems from the additional information provided by the NIR channel. We collected items requiring fine-grained perception, such as watch screwdrivers, fingerprints, grids, and color-coded resistors, to validate the proposed method's enhancement at the microscopic level, as shown in Fig. \ref{fig4}. Since the background normals of tactile surfaces are relatively easy to fit and learn, the focus was on the accuracy of the contact regions. Compared to the RGB-only results, the watch screwdriver exhibited clearer normal variations under NIR enhancement. The RGB-NIR approach successfully restored the microscopic structures of fingerprints, whereas the RGB-only method lost significant regions. For small items like color-coded resistors (approximately 3.5×2.5 mm), the RGB-only results appeared blurred in shape, while our method accurately estimated the geometric features of the resistors.

In summary, the PSNN demonstrates significant advantages over the look-up table method in normal estimation for curved surface sensors. The incorporation of NIR enhances the detail of the normal map, and the gradient estimation error is reduced by 40$\%$.

\subsection{Normal Integration by Depth Prior} 
We also conducted qualitative and quantitative experiments on normal integration.

Quantitatively, our method incorporates boundary constraints, enabling accurate integration on curved surfaces and improving performance, as shown in Tab. \ref{tab3}. Compared to the Fast Poisson solver commonly used for planar VTSs, the depth error generated by our method significantly decreased from 0.579 mm to 0.1255 mm. This is because the classical Fast Poisson solver assumes planar boundary conditions, which are unsuitable for curved surface reconstruction. In this paper, the edges of the sensor are assumed to undergo movement by default. Therefore, the depth information of the 10-pixel-wide image edges, calculated from the CAD model, is utilized as a prior for normal integration. By incorporating depth-based prior boundary conditions, we further enhanced the performance, reducing the depth error to 0.0406 mm.

Qualitatively, our normal integration results align with the normal estimation results, demonstrating that RGB-NIR enhances the fine-grained tactile perception performance of the VTS on curved surfaces. Compared to RGB-only results, the dense point cloud of the watch screwdriver under NIR enhancement exhibits clearer geometric contours. The microscopic structures of the fingerprints are fully restored, whereas the RGB-only approach loses details and appears blurred. In the RGB-NIR results, the relief effect of the point cloud of the grid pen holder is more pronounced, and the structure of the color-coded resistor is accurately reconstructed.

In summary, compared to the Fast Poisson method, our normal integration approach accurately generates point clouds and, with the addition of NIR, restores more faithful contour details.

\section{Conclusion \& Future Work}
\label{sec:conclusion}

This study introduced GelSplitter3D, a novel sensor system integrating RGB and NIR imaging, to enhance tactile perception on curved surfaces. Our PSNN and depth-constrained normal integration method significantly improved normal estimation and 3D reconstruction accuracy. The inclusion of NIR imaging enhanced fine-grained tactile perception, enabling clearer and more detailed reconstructions of complex objects. These advancements demonstrate the potential for more dexterous and human-like robotic grippers, improving performance in grasping and manipulation tasks.

\textbf{Future work} will focus on (i) optimizing the sensor structure to reduce its physical dimensions and enhance integration adaptability in confined spaces; (ii) implementing hardware-accelerated surface normal inference and integral algorithms on FPGA to enhance system response frequency; (iii) conducting grasping experiments using three-fingered flexible grippers to quantitatively evaluate the improvement in success rate through fine tactile feedback; (iv) migrating the proposed curved-surface vision tactile sensing framework to dexterous five-fingered hands, developing a vision-tactile fused perception system by integrating visual sensing and vision tactile feedback.

\textbf{Acknowledgement:} This work is supported by the Joint Funds of the National Natural Science Foundation of China (Grant No. U222A20208), the Natural Science Foundation Innovation Group Project of Hubei Province (Grant No. 2022CFA018), and the Key Research and Development Program of Guangdong Province (Grant No. 2022B0202010001-2).

{
\small
\setstretch{0.95}
\bibliographystyle{ieeetr}
\bibliography{reference}

\begin{thebibliography}{10}

\bibitem{wu2023learning}
Y.-H. Wu, J.~Wang, and X.~Wang, ``Learning generalizable dexterous manipulation from human grasp affordance,'' in {\em Conference on Robot Learning}, pp.~618--629, PMLR, 2023.

\bibitem{Sun2022}
H.~Sun, K.~J. Kuchenbecker, and G.~Martius, ``A soft thumb-sized vision-based sensor with accurate all-round force perception,'' {\em Nature Machine Intelligence}, vol.~4, pp.~135--145, 2 2022.

\bibitem{10563188}
S.~Li, Z.~Wang, C.~Wu, X.~Li, S.~Luo, B.~Fang, F.~Sun, X.-P. Zhang, and W.~Ding, ``When vision meets touch: A contemporary review for visuotactile sensors from the signal processing perspective,'' {\em IEEE Journal of Selected Topics in Signal Processing}, vol.~18, no.~3, pp.~267--287, 2024.

\bibitem{li2020skin}
G.~Li, S.~Liu, L.~Wang, and R.~Zhu, ``Skin-inspired quadruple tactile sensors integrated on a robot hand enable object recognition,'' {\em Science Robotics}, vol.~5, no.~49, p.~eabc8134, 2020.

\bibitem{sundaralingam2019robust}
B.~Sundaralingam, A.~S. Lambert, A.~Handa, B.~Boots, T.~Hermans, S.~Birchfield, N.~Ratliff, and D.~Fox, ``Robust learning of tactile force estimation through robot interaction,'' in {\em 2019 International Conference on Robotics and Automation (ICRA)}, pp.~9035--9042, IEEE, 2019.

\bibitem{babin2021mechanisms}
V.~Babin and C.~Gosselin, ``Mechanisms for robotic grasping and manipulation,'' {\em Annual Review of Control, Robotics, and Autonomous Systems}, vol.~4, no.~1, pp.~573--593, 2021.

\bibitem{wu2019tactile}
X.~A. Wu, T.~M. Huh, A.~Sabin, S.~A. Suresh, and M.~R. Cutkosky, ``Tactile sensing and terrain-based gait control for small legged robots,'' {\em IEEE Transactions on Robotics}, vol.~36, no.~1, pp.~15--27, 2019.

\bibitem{fang2019survey}
B.~Fang, S.~Jia, D.~Guo, M.~Xu, S.~Wen, and F.~Sun, ``Survey of imitation learning for robotic manipulation,'' {\em International Journal of Intelligent Robotics and Applications}, vol.~3, no.~4, pp.~362--369, 2019.

\bibitem{zhang2025artificial}
S.~Zhang, Y.~Yang, Y.~Sun, N.~Liu, F.~Sun, and B.~Fang, ``Artificial skin based on visuo-tactile sensing for 3d shape reconstruction: Material, method, and evaluation,'' {\em Advanced Functional Materials}, vol.~35, no.~1, p.~2411686, 2025.

\bibitem{romero2020soft}
B.~Romero, F.~Veiga, and E.~Adelson, ``Soft, round, high resolution tactile fingertip sensors for dexterous robotic manipulation,'' in {\em 2020 IEEE International Conference on Robotics and Automation (ICRA)}, pp.~4796--4802, IEEE, 2020.

\bibitem{do2022densetact}
W.~K. Do and M.~Kennedy, ``Densetact: Optical tactile sensor for dense shape reconstruction,'' in {\em 2022 International Conference on Robotics and Automation (ICRA)}, pp.~6188--6194, IEEE, 2022.

\bibitem{yuan2017gelsight}
W.~Yuan, S.~Dong, and E.~H. Adelson, ``Gelsight: High-resolution robot tactile sensors for estimating geometry and force,'' {\em Sensors}, vol.~17, no.~12, p.~2762, 2017.

\bibitem{gomes2020geltip}
D.~F. Gomes, Z.~Lin, and S.~Luo, ``Geltip: A finger-shaped optical tactile sensor for robotic manipulation,'' in {\em 2020 IEEE/RSJ International Conference on Intelligent Robots and Systems (IROS)}, pp.~9903--9909, IEEE, 2020.

\bibitem{tippur2023gelsight360}
M.~H. Tippur and E.~H. Adelson, ``Gelsight360: An omnidirectional camera-based tactile sensor for dexterous robotic manipulation,'' in {\em 2023 IEEE International Conference on Soft Robotics (RoboSoft)}, pp.~1--8, IEEE, 2023.

\bibitem{zhang2024gelroller}
Z.~Zhang, H.~Ma, Y.~Zhou, J.~Ji, and H.~Yang, ``Gelroller: A rolling vision-based tactile sensor for large surface reconstruction using self-supervised photometric stereo method,'' in {\em 2024 IEEE International Conference on Robotics and Automation (ICRA)}, pp.~7961--7967, IEEE, 2024.

\bibitem{tippur2024rainbowsight}
M.~H. Tippur and E.~H. Adelson, ``Rainbowsight: A family of generalizable, curved, camera-based tactile sensors for shape reconstruction,'' in {\em 2024 IEEE International Conference on Robotics and Automation (ICRA)}, pp.~1114--1120, IEEE, 2024.

\bibitem{10.1007/978-981-99-6498-7_2}
Y.~Lin, Y.~Zhou, K.~Huang, Q.~Zhong, T.~Cheng, H.~Yang, and Z.~Yin, ``Gelsplitter: Tactile reconstruction from near infrared and visible images,'' in {\em Intelligent Robotics and Applications}, pp.~14--25, Springer Nature Singapore, 2023.

\bibitem{lin20239dtact}
C.~Lin, H.~Zhang, J.~Xu, L.~Wu, and H.~Xu, ``9dtact: A compact vision-based tactile sensor for accurate 3d shape reconstruction and generalizable 6d force estimation,'' {\em IEEE Robotics and Automation Letters (RA-L)}, 2023.

\bibitem{xu2024dtactive}
J.~Xu, L.~Wu, C.~Lin, D.~Zhao, and H.~Xu, ``Dtactive: A vision-based tactile sensor with active surface,'' 2024.

\bibitem{liu2023gelsight}
S.~Q. Liu, L.~Z. Ya{\~n}ez, and E.~H. Adelson, ``Gelsight endoflex: A soft endoskeleton hand with continuous high-resolution tactile sensing,'' in {\em 2023 IEEE International Conference on Soft Robotics (RoboSoft)}, pp.~1--6, IEEE, 2023.

\bibitem{alspach2019soft}
A.~Alspach, K.~Hashimoto, N.~Kuppuswamy, and R.~Tedrake, ``Soft-bubble: A highly compliant dense geometry tactile sensor for robot manipulation,'' in {\em IEEE International Conference on Soft Robotics (RoboSoft)}, 2019.

\bibitem{zhao2023gelsight}
J.~Zhao and E.~H. Adelson, ``Gelsight svelte: A human finger-shaped single-camera tactile robot finger with large sensing coverage and proprioceptive sensing,'' in {\em 2023 IEEE/RSJ International Conference on Intelligent Robots and Systems (IROS)}, pp.~8979--8984, IEEE, 2023.

\bibitem{10035880}
J.~Hu, S.~Cui, S.~Wang, C.~Zhang, R.~Wang, L.~Chen, and Y.~Li, ``Gelstereo palm: A novel curved visuotactile sensor for 3-d geometry sensing,'' {\em IEEE Transactions on Industrial Informatics}, vol.~19, no.~11, pp.~10853--10863, 2023.

\bibitem{10358360}
S.~Cui, S.~Wang, C.~Zhang, R.~Wang, B.~Zhang, S.~Zhang, and Y.~Wang, ``Gelstereo biotip: Self-calibrating bionic fingertip visuotactile sensor for robotic manipulation,'' {\em IEEE/ASME Transactions on Mechatronics}, vol.~29, no.~4, pp.~2451--2462, 2024.

\bibitem{sun2022soft}
H.~Sun, K.~J. Kuchenbecker, and G.~Martius, ``A soft thumb-sized vision-based sensor with accurate all-round force perception,'' {\em Nature Machine Intelligence}, vol.~4, no.~2, pp.~135--145, 2022.

\bibitem{cao2022bilateral}
X.~Cao, H.~Santo, B.~Shi, F.~Okura, and Y.~Matsushita, ``Bilateral normal integration,'' in {\em European Conference on Computer Vision}, pp.~552--567, Springer, 2022.

\bibitem{xiu2023econ}
Y.~Xiu, J.~Yang, X.~Cao, D.~Tzionas, and M.~J. Black, ``Econ: Explicit clothed humans optimized via normal integration,'' in {\em Proceedings of the IEEE/CVF conference on computer vision and pattern recognition}, pp.~512--523, 2023.

\bibitem{cao2024supernormal}
X.~Cao and T.~Taketomi, ``Supernormal: Neural surface reconstruction via multi-view normal integration,'' in {\em Proceedings of the IEEE/CVF Conference on Computer Vision and Pattern Recognition}, pp.~20581--20590, 2024.

\bibitem{padmanabha2020omnitact}
A.~Padmanabha, F.~Ebert, S.~Tian, R.~Calandra, C.~Finn, and S.~Levine, ``Omnitact: A multi-directional high-resolution touch sensor,'' in {\em IEEE International Conference on Robotics and Automation (ICRA)}, 2020.

\bibitem{woodham1979photometric}
R.~J. Woodham, ``Photometric stereo: A reflectance map technique for determining surface orientation from image intensity,'' in {\em Image understanding systems and industrial applications I}, vol.~155, pp.~136--143, SPIE, 1979.

\bibitem{kazhdan2006poisson}
M.~Kazhdan, M.~Bolitho, and H.~Hoppe, ``Poisson surface reconstruction,'' in {\em Proceedings of the fourth Eurographics symposium on Geometry processing}, vol.~7, 2006.

\end{thebibliography}
}
\end{document}